%% file: main.tex
\documentclass[10pt,twocolumn,letterpaper]{article}

\usepackage[pagenumbers]{cvpr}   
\usepackage{algorithm}
\usepackage{algorithmic}
\usepackage[percent]{overpic}
\usepackage{microtype}
\usepackage{setspace}

\usepackage{booktabs}
\usepackage{multirow}
\usepackage{amssymb}
\usepackage{bbding}

\usepackage[numbers]{natbib}


\definecolor{cvprblue}{rgb}{0.21,0.49,0.74}
\usepackage[pagebackref,breaklinks,colorlinks,allcolors=cvprblue]{hyperref}

\def\paperID{5467} 
\def\confName{CVPR}
\def\confYear{2025}

\title{COB-GS: Clear Object Boundaries in 3DGS Segmentation Based on Boundary-Adaptive Gaussian Splitting}

\author{
    {Jiaxin Zhang, Junjun Jiang\thanks{Correspondence to: Junjun Jiang (junjunjiang@hit.edu.cn)} , Youyu Chen, Kui Jiang, Xianming Liu}\\
    {Harbin Institute of Technology}\\
}

\begin{document}
\maketitle
\input{sec/0_abstract}    
\input{sec/1_intro}
\input{sec/2_related}
\input{sec/3_method}
\input{sec/4_experiment}
\input{sec/5_discussion_and_conclusion}

{
    \small
    \bibliographystyle{ieeenat_fullname}
    \bibliography{main}
}

\input{sec/X_suppl}

\end{document}

%% file: sec/0_abstract.tex
\begin{abstract}

Accurate object segmentation is crucial for high-quality scene understanding in the 3D vision domain. However, 3D segmentation based on 3D Gaussian Splatting (3DGS) struggles with accurately delineating object boundaries, as Gaussian primitives often span across object edges due to their inherent volume and the lack of semantic guidance during training. In order to tackle these challenges, we introduce Clear Object Boundaries for 3DGS Segmentation (COB-GS), which aims to improve segmentation accuracy by clearly delineating blurry boundaries of interwoven Gaussian primitives within the scene. Unlike existing approaches that remove ambiguous Gaussians and sacrifice visual quality, COB-GS, as a 3DGS refinement method, jointly optimizes semantic and visual information, allowing the two different levels to cooperate with each other effectively. Specifically, for the semantic guidance, we introduce a boundary-adaptive Gaussian splitting technique that leverages semantic gradient statistics to identify and split ambiguous Gaussians, aligning them closely with object boundaries. For the visual optimization, we rectify the degraded suboptimal texture of the 3DGS scene, particularly along the refined boundary structures. Experimental results show that COB-GS substantially improves segmentation accuracy and robustness against inaccurate masks from pre-trained model, yielding clear boundaries while preserving high visual quality. Code is available at \url{https://github.com/ZestfulJX/COB-GS}. 
\end{abstract}

%% file: sec/1_intro.tex
\vspace{-0.45cm}
\section{Introduction}
\label{sec:intro}
\vspace{-0.1cm}
\begin{spacing}{0.99} 
Understanding and interacting with 3D scenes has long been a critical challenge in computer vision and computer graphics. This task involves reconstructing 3D scenes from collections of images or videos, as well as accurately perceiving and segmenting 3D structures. In recent years, researchers have conducted extensive research on 3D scene representation and perception. Among these advancements, 3D Gaussian Splatting (3DGS)~\cite{kerbl20233d}, an emerging real-time radiance field rendering technique, has demonstrated comparable novel view rendering quality against Neural Radiance Fields (NeRF)~\cite{mildenhall2020nerf}, while significantly outperforming NeRF in the speed of optimization and rendering, enabling real-time rendering. As an explicit scene representation, 3DGS opens new avenues for 3D scene perception and interaction.
\end{spacing}
\begin{figure}[t]
    \centering
    \vspace{-0.2cm}
    \begin{overpic}[width=1.0\linewidth]{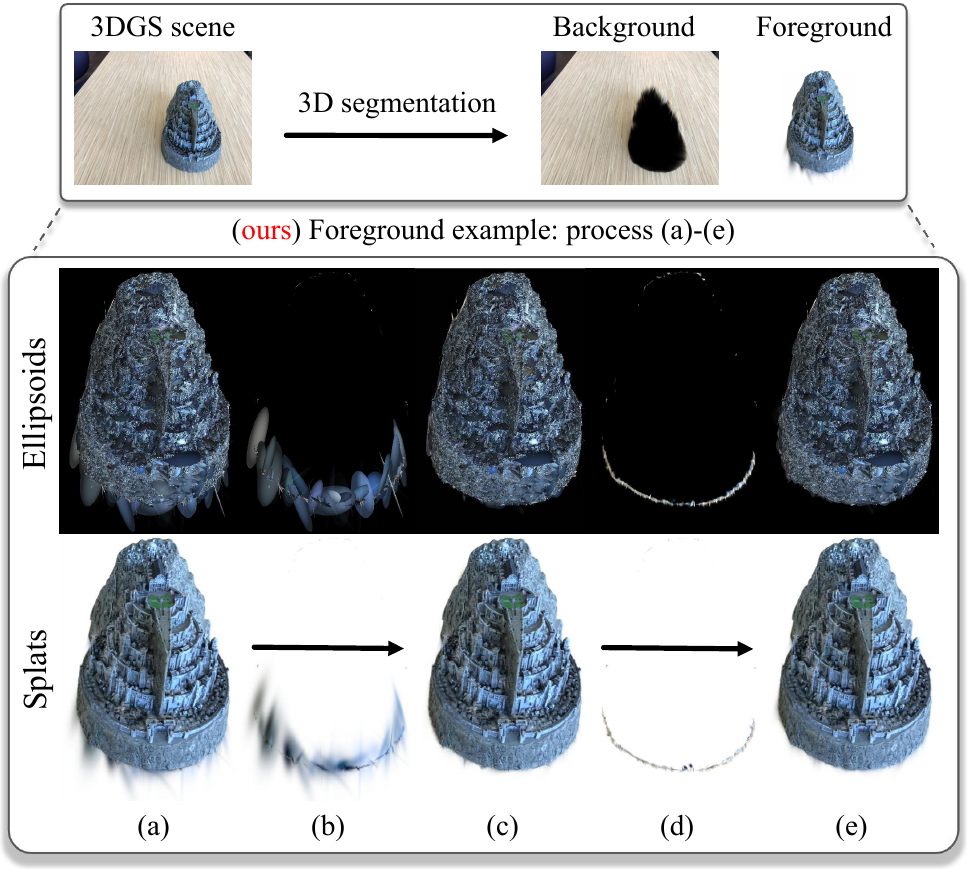}
        \put(28, 29.5){\scriptsize {\cref{method:Localization and Splitting}}}
        \put(28, 25.5){\scriptsize {\cref{method:texture} }}
        \put(64, 27.5){\scriptsize {\cref{method:robustness} }}
    \end{overpic}
    \captionsetup{singlelinecheck=false}

    \caption{
        Problems on 3DGS segmentation. 
        Our task is to achieve high-quality 3D segmentation on 3DGS, including foreground and background. Take foreground as an example: (a) unclear segmentation result of existing methods; (b) blurry boundary Gaussians; (c) segmentation results from joint optimization; (d) tiny ambiguous Gaussians due to incorrect masks from pre-trained models; (e) final segmentation results after improving robustness from COB-GS.
    }
    \label{fig:teaser}
    \vspace{-0.3cm}
\end{figure}

\begin{spacing}{0.99} 
3D segmentation is fundamental for effective 3D scene perception and interaction, and it is crucial to obtain accurate segmentation results within the neural 3D representations exemplified by 3DGS. 
Currently, two predominant methodologies exist for executing the segmentation of 3DGS: feature-based methods and mask-based methods. 
Feature-based methods typically operate in conjunction with 
3D scene reconstruction to learn distinctive feature properties for each Gaussian primitive. 
During the segmentation phase, the similarity between the 3D Gaussian features and the queried feature is computed to specify the Gaussian primitives with the desired semantic~\cite{shi2024language, zhou2024feature, qin2023langsplat, gaussian_grouping, cen2023saga}. 
However, these methods encounter challenges for inefficient training and rendering processes, as well as the inherent ambiguity associated with high-dimensional feature representations.
\end{spacing}

\begin{spacing}{0.98} 
To mitigate these concerns, mask-based post-processing methods leverage the semantic masks of input views from Segment Anything Model (SAM)~\cite{kirillov2023segany} to learn category labels for each 3D Gaussian in the reconstructed 3DGS scene, filtering these Gaussian primitives with the specified query label to perform 3D segmentation~\cite{cen2023segment, hu2024semantic, flashsplat}. 
Despite this advancement, the original scene reconstruction often neglects the semantic information, focusing primarily on visual optimization while overlooking the volumetric characteristics of the Gaussian primitives. 
This oversight will lead to blurred labels for the boundary Gaussians during scene segmentation, which result in imprecise segmentation results characterized by blurry object edges, as illustrated in Figure \ref{fig:teaser}. 
Some of existing methods delete the ambiguous boundary Gaussians directly~\cite{cen2023segment, flashsplat}. 
However, simply removing the Gaussian primitives on the boundary will disturb the visual quality.

To confront these issues, we propose COB-GS, a 3DGS refinement and segmentation method that that jointly optimizes semantics and appearance to register semantic masks to Gaussian primitives. 
Similar to existing approaches, we introduce $mask$ $label$ as an additional attribute to each Gaussian for segmentation. 
Moreover, we reveal a strong correlation between the gradient direction of these labels and the supervising category at the pixel level, which is a strong discriminator of ambiguous Gaussian primitives on the boundary. 
Specifically, during the mask optimization phase, our approach utilizes gradient statistics of the mask label to identify and split boundary Gaussians, allowing precise alignment with object edges. In the scene optimization phase, we refine the scene texture on the correct boundary structure to maintain visual quality. 
After scene optimization, 3D segmentation focuses on filtering the mask labels. 
Additionally, we distinguish between boundary blurring due to the volume of Gaussians and inaccurate masks. By refining tiny boundary Gaussians, we exclusively enhance the robustness of our method against inaccurate masks from the pre-trained model.
Finally, we introduce a two-stage mask generation method based on SAM2~\cite{ravi2024sam2}, significantly simplifying the extraction of region-of-interest masks from 3D reconstruction datasets.
\end{spacing}
To summarize our contributions in a few words:
\begin{itemize}[leftmargin=*]
    \item To the best of our knowledge, we are the first 3DGS segmentation method explicitly designed to jointly optimize semantic and visual information, ensuring they enhance one another, aligning Gaussians with object edges to efficiently obtain clear boundaries and improve visual quality.
    \item We propose a boundary-adaptive Gaussian splitting technique that leverages gradient information from semantics to refine ambiguous boundary Gaussians, along with a boundary-guided scene texture restoration method to preserve scene visual quality on the refined boundary.
    \item We demonstrate the robustness of our method to inaccurate masks from pre-trained model by extracting and refining the tiny boundary Gaussians during optimization.
    \item We introduce a two-stage mask generation method using SAM2 based on text prompts, effectively addressing the object continuity issues in long sequence prediction.
\end{itemize}

%% file: sec/2_related.tex
\vspace{-0.2cm}
\section{Related Works}
\label{sec:related_works}
\textbf{3D Gaussian Splatting.} As an emerging real-time inverse rendering technology, 3DGS~\cite{kerbl20233d} has been proven to match the high rendering quality of the NeRF~\cite{mildenhall2020nerf} in the novel view synthesis, and the speed is much faster than NeRF. Current advancements in 3DGS focus on improving aspects such as reconstruction quality~\cite{Yu2024MipSplatting,scaffoldgs,cheng2024gaussianpro}, reconstruction speed~\cite{taming3dgs,fang2024minisplattingrepresentingscenesconstrained}, and storage consumption~\cite{fan2023lightgaussian,lee2024c3dgs}. Other efforts aim to address special cases, including dynamic scenes~\cite{liang2023gaufregaussiandeformationfields,Wu_2024_CVPR} and challenging inputs~\cite{zhu2023FSGS,charatan23pixelsplat}. As an explicit representation, 3DGS also provides more possibilities for 3D editing~\cite{chen2023gaussianeditor,xie2023physgaussian,qiu-2024-featuresplatting} and 3D generation~\cite{tang2023dreamgaussian,EnVision2023luciddreamer}. Our approach focuses on the 3D segmentation with clear boundaries within 3DGS scenes. 

\noindent\textbf{3D Neural Scene Segmentation.} Recent advancements in neural 3D scene representation~\cite{mildenhall2020nerf,Sun_2022_CVPR,Fridovich-Keil_2022_CVPR,kerbl20233d} and 2D foundational models~\cite{kirillov2023segany,ravi2024sam2,radford2021learningtransferablevisualmodels,caron2021emerging,depthanything} have significantly enhanced the ability to perceive and interact with 3D scenes. These methods focus on learning additional attributes for 3D representations leveraging foundational 2D models, expanding beyond color to address a range of tasks. One key area of research is 3D segmentation. Early NeRF~\cite{mildenhall2020nerf}, as an implicit neural representation, was commonly used for 3D segmentation~\cite{kim2024garfieldgroupradiancefields,Zhi:etal:ICCV2021,liu2023weakly,lerf2023,isrfgoel2023,instructnerf2023,spinnerf,cen2023segment}. However, it faced challenges with decoupling due to the inherent limitations of neural networks. In contrast, 3DGS provides an explicit representation that facilitates better region decoupling in 3D segmentation, offering a more effective alternative to NeRF. 

There are two primary methods for 3D segmentation on 3DGS. The feature-based approach~\cite{qin2023langsplat,shi2024language,zhou2024feature,choi2025click}: Grouping Gaussian~\cite{gaussian_grouping} learns identity encodings for each 3D Gaussian and groups them with the same encodings, enforcing spatial consistency in 3D to constrain the identity encoding learning process. SAGA~\cite{cen2023saga} improves segmentation across different scales by incorporating scale-related affinity features into each Gaussian. The closest approach to ours is the mask-based approach~\cite{cen2023segment,hu2024semantic,flashsplat}, in which SAGD~\cite{hu2024semantic} employs a cross-view label voting mechanism and Gaussian decomposition to enhance foreground quality. FlashSplat~\cite{flashsplat} addresses the inverse rendering of 2D masks using linear programming, and introduces the background bias to eliminate boundary Gaussian. 
Existing methods typically separate scene segmentation from reconstruction, neglecting the volume of 3D Gaussians, leading to inaccurate segmentation and blurry boundaries. In contrast, our method uses joint optimization of semantics and texture to achieve clear object boundaries while preserving visual quality.

%% file: sec/3_method.tex
\begin{figure*}
    \centering
    \includegraphics[width=1.0\linewidth]{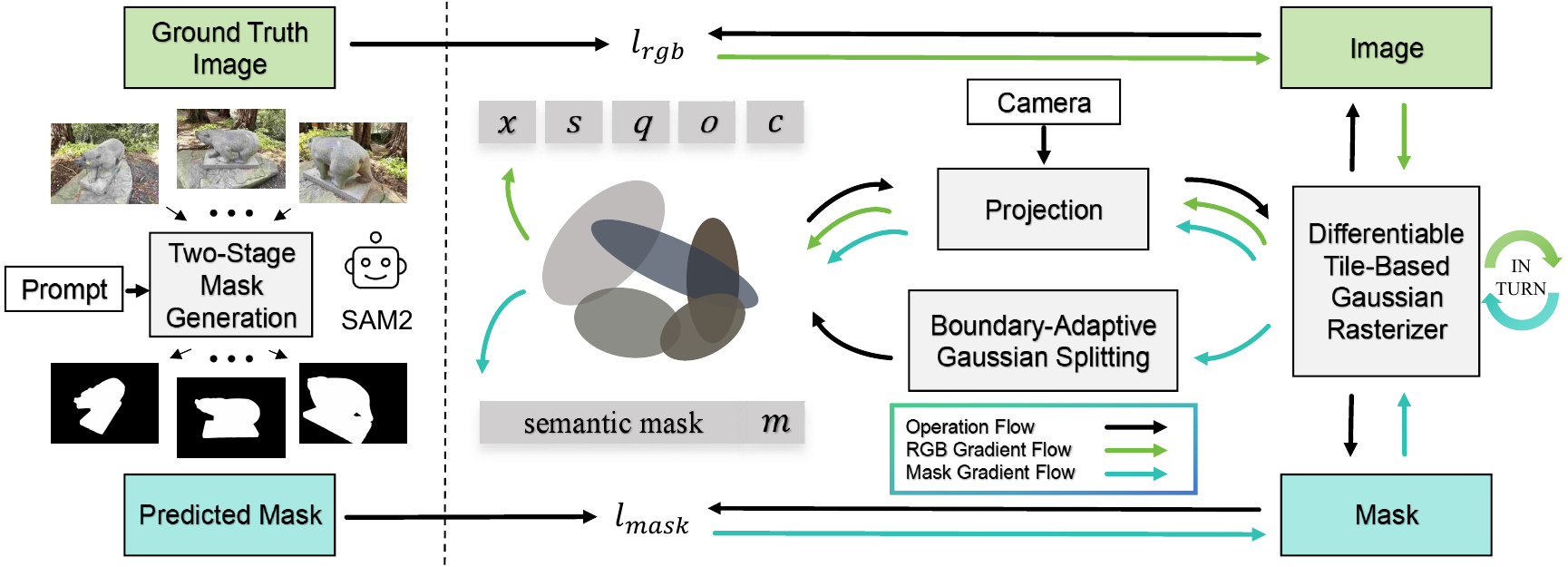}
    \vspace{-0.5cm}
    \captionsetup{singlelinecheck=false}
    \caption{
        Pipeline of our clear object boundary segmentation method for 3DGS. 
        On the left, we present our two-stage mask generation method, which utilizes SAM2 perform mask prediction on image sequences based on text prompt to obtain masks for regions of interest. Images and masks serve as supervision for 3DGS refinement. On the right, for the reconstructed 3DGS scene, we jointly and alternately optimize the mask and texture. For the mask optimization, boundary-adaptive Gaussian splitting is performed to refine boundary structure.
        }
    \label{fig:pipeline}
    \vspace{-0.25cm}
\end{figure*}

\section{Method}
\label{sec:method}
In the section, we first introduce the principles of 3D Gaussian Splatting in Sec \ref{method:Preliminary} to clarify the rasterization process. In Sec \ref{method:Mask Label Optimization Process}, we elaborate on how to  jointly optimize semantics and texture, and its core is the boundary-adaptive Gaussian splitting based on mask gradient statistics. Finally, we introduce two-stage mask generation on SAM2 in Sec \ref{method:mask}. Figure \ref{fig:pipeline} depicts the pipeline of our proposed COB-GS.

\subsection{Preliminary: 3D Gaussian Splatting}
\label{method:Preliminary}
3D Gaussian Splatting (3DGS) \cite{kerbl20233d} is an emerging real-time radiance field rendering technique that achieves rendering quality comparable to NeRF. As an explicit scene representation, it opens new possibilities for scene segmentation and editing.
Given a set of $V$ input views $\{I^v\}$ along with their corresponding camera poses, 3DGS can represent the scene by learning a set of Gaussians $\{G_i\}$. In the original 3DGS, the $i$-th 3D Gaussian primitive can be parameterized as $G_i = \{x_i, s_i, q_i, o_i, c_i\}$, where $x_i \in \mathbb{R}^3$ is the center position, $s_i \in \mathbb{R}^3$ is the scale, $q_i \in \mathbb{R}^4$ is the rotation, $o_i \in \mathbb{R}$ is the opacity, and $c_i \in \mathbb{R}^{48}$ denotes the color.

During rendering, all 3D Gaussians are first projected onto the image plane as 2D Gaussians. The set of projected 3D Gaussians in the shared space is then accessed in parallel in the form of pixel blocks. Specifically, when rendering a pixel, traditional alpha compositing blends the target attributes \( p_i \) (color, depth, or semantic features) of the 2D Gaussians into the pixel space \( P \):

\[
P = \sum_{i=1}^{N} p_i \alpha_i \prod_{j=1}^{i-1} (1 - \alpha_j) = \sum_{i=1}^{N} p_i \alpha_i T_i, \tag{1}
\]
where \( \alpha_i \) is the product of the opacity of the \( i \)-th Gaussian primitive and the probability of its projection's distance from the center position of the 2D Gaussians decaying exponentially. \( T_i = \prod_{j=1}^{i-1} (1 - \alpha_j) \) represents the transmittance, indicating the score of penetrating through the previous \( i-1 \) Gaussians to the current Gaussian.

\subsection{Boundary-Aware Object Segmentation}
\label{method:Mask Label Optimization Process}

For efficiency, we introduce a continuous $mask$ $label$ \( m_i \in (0, 1) \) for each Gaussian, where \( m_i \) close to 1 indicates that the \( i \)-th Gaussian is necessary for segmenting the 3D object, while \( m_i \) close to 0 indicates that the \( i \)-th Gaussian belongs to the background. 
Similar to the color rendering process, the mask labels of the 3D Gaussian primitives are combined through alpha compositing to yield the mask result in the 2D pixel space \( M_{\text{render}} \):

\[
M_{\text{render}} = \sum_{i=1}^{N} m_i \alpha_i \prod_{j=1}^{i-1} (1 - \alpha_j) = \sum_{i=1}^{N} m_i \alpha_i T_i. \tag{2}
\]

Inspired by SA3D~\cite{cen2023segment}, we use a similar loss function to supervise the mask label training process:

\[
\mathcal{L}_{\text{mask}} = \sum_{M_{jk}^v} M_{jk}^v \cdot M_{\text{render}}^v + \sum_{M_{jk}^v} (1 - M_{jk}^v) \cdot M_{\text{render}}^v, \tag{3}
\]
where \( M_{jk}^v \in \{0, 1\} \) is the ground truth mask at pixel location $(j,k)$ for view \( v \) from \( \{M^v\} \), and the mask generation method is detailed in Sec \ref{method:mask}. Unlike SA3D, which does not limit the range of $m_i$, we constrain the mask labels \( m_i \in (0, 1) \). Based on the absorption light score of all sampled Gaussians in the alpha compositing formula \( 0 < \sum_i \alpha_i T_i < 1 \), we can deduce \( 0 < \sum_i m_i \alpha_i T_i < 1 \), ensuring that \( \mathcal{L}_{\text{mask}} \) converges. Additionally, we eliminate hyperparameters that determine negative loss in SA3D and instead emphasize learning the background to achieve 3D segmentation rather than just foreground extraction.

\subsubsection{Boundary-Adaptive Gaussian Splitting}
\label{method:Localization and Splitting}

The original 3DGS relied on RGB supervision, which lacked the object-level semantic information used to shape 3D Gaussians. As a result, the segmentation of freezing geometry and texture will lead to semantically ambiguous boundary Gaussians. Therefore, it is crucial to locate and split these ambiguous Gaussians for obtaining clear object boundaries.

To address this issue, unlike the inefficient forward voting process used in existing methods~\cite{flashsplat,hu2024semantic} for each Gaussian, we use the gradient of mask labels in the mask optimization phase for ambiguous Gaussian identification. Specifically, for a pixel location $(j,k)$ at viewpoint \( v \), the derivative of \( \mathcal{L}_{\text{mask}}^{vjk} \) with respect to \( m_i \) can be expressed as:

\[
\frac{\mathrm{d} \mathcal{L}_{\text{mask}}^{vjk}}{\mathrm{d} m_i} = 
\begin{cases} 
- \alpha_i T_i, & \text{if } M_{jk}^v = 1 \\ 
\alpha_i T_i, & \text{if } M_{jk}^v = 0 
\end{cases} \tag{4}
\]

During optimization, the viewpoint \( v \) is used as the minimum unit of the gradient calculation, and Gaussian has the volume. Therefore the gradient calculation for \( \mathcal{L}_{\text{mask}}^v \) with respect to \( m_i \) is influenced by multiple pixels, leading to:

\[
\frac{\mathrm{d} \mathcal{L}_{\text{mask}}^v}{\mathrm{d} m_i} = \sum_{j=1}^{N_{v,i}^+} (-\alpha_j T_j) + \sum_{j=1}^{N_{v,i}^-} (\alpha_j T_j), \tag{5}
\]
where \( N_{v,i}^+ \) and \( N_{v,i}^- \) represent the number of signals supervising the \( i \)-th Gaussian with ground truth masks of 1 and 0, respectively. Thus, the cumulative gradient at a viewpoint \( v \) is not effective for distinguishing ambiguous boundary Gaussians, while the sign of the gradient under a pixel reflects the supervised category of the mask labels. 

Motivated by the relationship between gradient direction and supervising signals, we introduce a new variable for each Gaussian during backpropagation to capture the consistency strength of supervision signals on the mask label under a viewpoint, using the absolute value of the relative distance:

\[
mask\_sig_{v,i} = \left| \frac{N_{v,i}^+ - N_{v,i}^-}{N_{v,i}^+ + N_{v,i}^- + \epsilon} \right|, \tag{6}
\]
where \( \epsilon \) is a small constant. 
The closer $mask\_sig$ is to 0, the stronger the supervision by the inconsistent signal. 
During the optimization of mask labels, Gaussians with an absolute relative distance below a threshold are identified as the semantically ambiguous boundary Gaussian set \( \{G_i\}_B \).

\[
\{G_i\}_B = \{G_i \mid i \in \mathcal{I} \land \left( \frac{1}{V} \sum_{v=1}^{V} {mask\_sig}_{v,i} < \delta \right) \}, \tag{7}
\]
where \( \delta \) is the threshold and $\mathcal{I} = \{1, 2, 3, \ldots, |\{G_i\}|\}$. During the splitting process, we refer to the original 3D Gaussian Splatting. First, we exclude small-scale Gaussians from \( \{G_i\}_B \). For the remaining larger Gaussians, we replace each with two smaller Gaussians, scaling down from the original. We then use the original Gaussian as the probability density function (PDF) to sample their initial positions.

\begin{figure}
    \centering
    \includegraphics[width=1.0\linewidth]{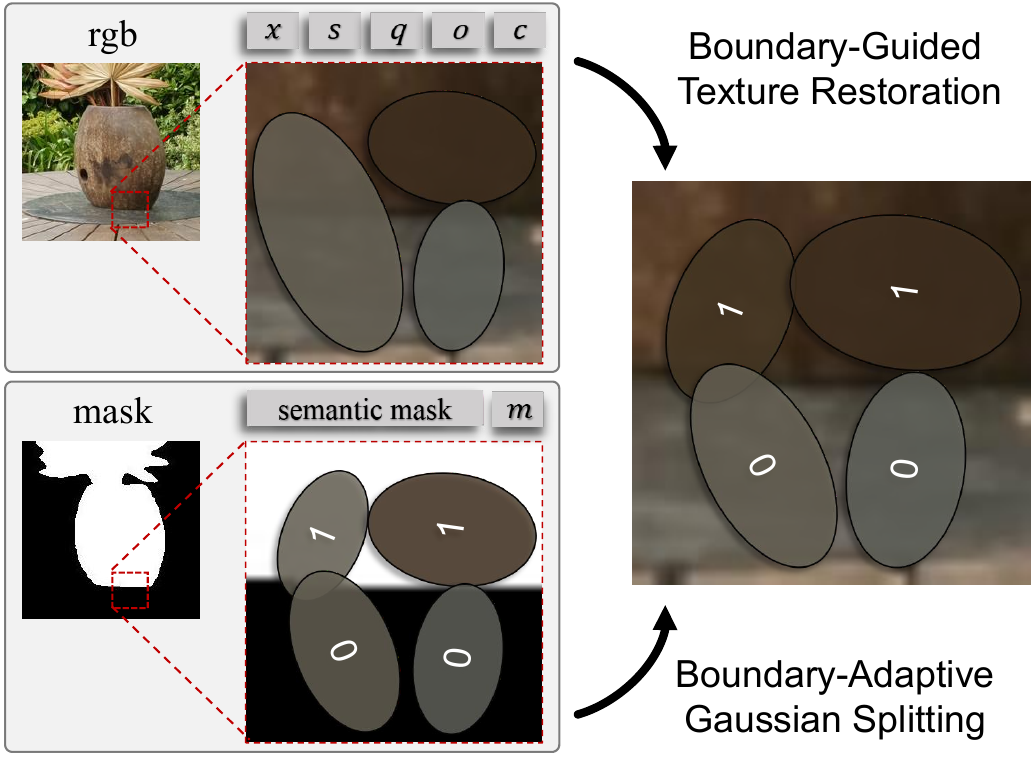}
    \captionsetup{singlelinecheck=false}
    \caption{
        Supervision based solely on texture results in large Gaussians due to the similarity in textures between objects. However, object-level mask supervision facilitates the differentiation of object edges. This allows 3D Gaussians to split along object edges, while also guiding the correct restoration of scene textures.
    }
    \label{fig:mask+rgb}
    \vspace{-5pt}
\end{figure}

\begin{figure*}
    \centering
    \includegraphics[width=1.0\linewidth]{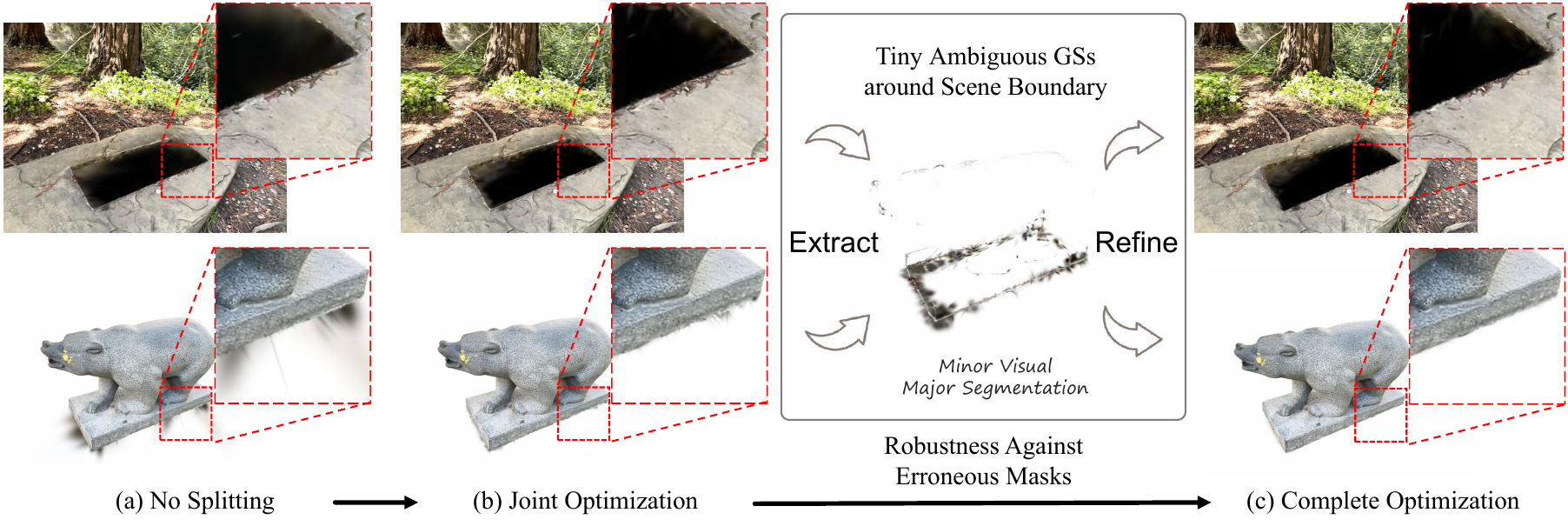}
    \vspace{-15pt}
    \captionsetup{singlelinecheck=false}
    \caption{
        Visualization of different processing phases. 
        (a) Optimizing the mask labels without Gaussian splitting results in unclear boundary segmentation. (b) Jointly optimizing masks and textures with boundary-adaptive Gaussian splitting effectively reduces large ambiguous Gaussians while still leaving tiny ones at the boundaries. (c) Extract and refine tiny ambiguous boundary Gaussians obtained by non-convergent splitting. These tiny Gaussians have little impact on visual quality but affect boundary clarity in 3D segmentation.
    }
    \label{fig:fine gs}
    \vspace{-5pt}
\end{figure*}

\subsubsection{Boundary-Guided Scene Texture Restoration}
\label{method:texture}

Existing scene segmentation methods directly remove boundary ambiguous Gaussians~\cite{cen2023segment, flashsplat} or focus only on the scales of foreground Gaussians~\cite{hu2024semantic}. These rough methods can compromise visual quality, and object-level semantic conditions are not fully utilized in scene texture learning.

To solve this problem, our unique insight is that accurate object boundaries can enhance 3D segmentation and improve scene texture optimization. As shown in Figure \ref{fig:mask+rgb}, we alternately learn the mask labels and the geometric and texture of the Gaussians. Incorporating object-level semantic information effectively limits the volume of boundary Gaussians, and optimizing texture on accurate boundary structure enhances visual quality for new views.

Specifically, the loss function for learning the geometric and texture information of the Gaussians is consistent with the original Gaussian optimization process:

\[
\mathcal{L}_{\text{rgb}} = (1 - \lambda) \mathcal{L}_1 + \lambda \mathcal{L}_{\text{D-SSIM}}, \tag{8}
\]
where \( \lambda \) is a hyperparameter. In the alternating optimization process, we first optimize the mask labels by minimizing \( \mathcal{L}_{\text{mask}} \) while freezing the geometry and texture of the Gaussians. As mentioned in Sec  \ref{method:Localization and Splitting}, we locate and split the semantically ambiguous boundary Gaussians under a certain number of training views. During this phase, the geometry and texture of the scene are compromised, leading to a significant degradation in visual quality. Therefore, we optimize \( \mathcal{L}_{\text{rgb}} \), freezing the mask labels to refine the scene's geometry and texture details on the accurate boundary structure. The above two stages are performed iteratively and alternately to fuse the information of two modalities to refine 3DGS scene representation, thereby ensuring precise object boundaries and maintaining the visual quality of viewpoint synthesis.

\subsubsection{Robustness Against Erroneous Masks}
\label{method:robustness}
Through alternating optimization of masks and textures, the number of semantically ambiguous Gaussians should progressively decline. In the segmentation phase, we segment the scene based on mask labels. However, experiments reveal that although the quantity of large ambiguous Gaussians reduces, numerous small ambiguous ones remain. 

In fact, the binary masks \( \{M^v\} \) predicted by the trained 2D vision model exhibit discreteness, which often leads to inaccuracies and inconsistencies in object boundary predictions across different views. This limitation may prevent mask labels of the boundary Gaussians from converging during the optimization process.
In contrast to existing methods~\cite{flashsplat} that roughly address boundary blur caused by inaccurate masks and Gaussian volumes in a combined manner, our approach utilizes the final stage of joint optimization to exclusively enhance robustness against inaccurate masks.
Specifically, we identify the tiny boundary Gaussian with ambiguous semantics based on lower values of $mask\_sig$ and scale $s$, and the visualization result is shown in Figure \ref{fig:fine gs}. While these Gaussians minimally affect scene visual quality without 3D segmentation, their removal is essential for achieving clear and complete boundaries for both foreground objects and the background in the 3D segmentation process.
\subsubsection{Multi-Object 3D Segmentation}
\label{method:Multi-Object}
Real 3DGS scene contains multiple objects. Feature-based methods typically require a time-consuming full-scene training process to assign features to each Gaussian for querying. These methods use masks to regularize scene for accurate object boundaries, which limits the granularity of segmentation. For instance, the requirement of clear boundary segmentation of two granularities, \textit{``bear's head''} and \textit{``bear''}, is hard to meet in a trained scene. Similarly, mask-based methods for learning full-scene labels face comparable challenges.

We propose decomposing multi-object segmentation into sequential single-object 3D segmentation. By adding a single integer storage to each Gaussian and utilizing rasterized real-time rendering, we accelerate individual splits, addressing granularity issues. Specifically, for \( K \) objects, we define a mask set \( \{M^v\}_k \) where \( k \in \{1, 2, \ldots, K\} \). When optimizing for the \( k \)-th object, we perform Gaussian splitting, jointly optimizing with texture to obtain a new 3DGS \( \{G_i\}_k \) with clear object boundaries. The optimization process for subsequent objects is conducted on the updated 3DGS \( \{G_i\}_k \), iterating until all \( K \) objects are optimized. For the specific object to be segmented, a single round of rapid mask label learning is sufficient to achieve clear segmentation results.
\subsection{Two-Stage Mask Generation}
\label{method:mask}
Mask-based 3DGS segmentation involves generating masks for the target objects based on input images. In our approach, the supervision data consists of \( V \) input images \( \{I^v\} \) paired with corresponding 2D binary masks \( \{M^v\} \). With the development of foundational models such as SAM2~\cite{ravi2024sam2}, mask prediction across image sequences has become viable, significantly improving inter-frame consistency. However, SAM2 encounters challenges with object continuity over long sequences, potentially failing to infer objects that are heavily occluded due to discontinuities in visual information.

To address this limitation, we propose a two-stage mask generation approach utilizing text prompts. In the coarse stage, Grounding-DINO~\cite{caron2021emerging} is used to extract box prompts from the frame with lower text confidence, which are then applied across the entire sequence for initial mask predictions. 
In the fine-grained stage, we use Grounding-DINO with higher text confidence to extract box prompts for subsequences where mask prediction was disrupted in the coarse stage. 
These prompts generate the final mask predictions for the subsequence. See supplementary material for details.

%% file: sec/4_experiment.tex
\vspace{-5pt}
\section{Experiments}
\label{experiment}
We demonstrate the effectiveness of our method both quantitatively and qualitatively. For quantification, we utilize the NVOS dataset~\cite{ren-cvpr2022-nvos}, which is derived from the LLFF dataset and provides ground truth masks with precise object edges. For qualitative evaluation, we employ scenes from various datasets, including LLFF~\cite{mildenhall2019llff}, MIP-360~\cite{barron2021mipnerf}, T\&T~\cite{10.1145/3072959.3073599}, and LERF~\cite{lerf2023}. These datasets encompass real-world complex scenes, including indoor and outdoor scenes, as well as forward and surrounding scenes. Implementation details and more experiments are provided in supplementary material.

\subsection{Quantitative Results}
\label{method:Quantitative Results}

The NVOS dataset contains eight scenes. Each scene includes a reference view and a target view with a clear GT mask. Our method utilizes the annotated masks from the reference view to extract prompt points, which are then passed to the remaining views using SAM2 to generate masks. 

\noindent\textbf{Segmentation evaluation.} We extract a Gaussian set corresponding to the object and render the 2D mask for the target view. We then calculate the mean IoU and mean accuracy between the GT mask and the rendered mask. Results are shown in Table \ref{tab:nvos1}. The NeRF-based method exhibits limited segmentation detail due to inadequate scene representation. In 3DGS segmentation, feature-based approaches result in unsmooth object boundaries because of high-dimensional feature ambiguity. The most similar method to ours is the mask-based 3DGS method. SAGD~\cite{hu2024semantic} lacks robustness to erroneous GT masks, leading to small Gaussian artifacts along segmented edges. FlashSplat~\cite{flashsplat} reduces edge blur but compromises the structural integrity of objects. 

\noindent\textbf{Visual evaluation.} 
Due to the lack of reference images for object segmentation results, we use CLIP-IQA~\cite{wang2022exploring}, a no-reference IQA that evaluates how well an image matches the given text prompt.
We set three prompts focusing on boundary quality to comprehensively evaluate the segmentation results, as shown in Table \ref{tab:nvos2}. SAGD~\cite{hu2024semantic} and FlashSplat~\cite{flashsplat} process Gaussians roughly, destroying the appearance and sacrificing visual quality for segmentation accuracy. 
SA3D~\cite{cen2023segment} exhibits obvious ambiguous boundary Gaussians. Ours demonstrates high visual quality across the board.
\begin{table}[htbp]
    \centering
    \caption{Quantitative segmentation results on NVOS dataset.}
    \label{tab:nvos1}
    \small
    \begin{tabular}{llcc}
    \toprule
    Category   & Method     & mIoU (\%)& mAcc (\%) \\
    \midrule
    \multirow{4}{*}{NeRF-based} & NVOS~\cite{ren-cvpr2022-nvos} & 70.1  & 92.0 \\
               & ISRF~\cite{isrfgoel2023} & 83.8 & 96.4 \\
               & SA3D~\cite{cen2023segment} & 90.3 & 98.2  \\
               & OmniSeg3D~\cite{ying2023omniseg3d} & 91.7 & 98.4  \\
    \midrule
    \multirow{5}{*}{3DGS-based} & SAGD~\cite{hu2024semantic} & 90.4 & 98.2  \\
               & SA3D-GS~\cite{cen2023segment} & 90.7 & 98.3  \\
               & SAGA~\cite{cen2023saga} & 90.9 & 98.3  \\
               & FlashSplat~\cite{flashsplat} & 91.8 & \textbf{98.6}  \\
               & COB-GS (ours) & \textbf{92.1} & \textbf{98.6} \\
    \bottomrule
    \end{tabular}
    \vspace{-5pt}
\end{table}

\begin{table}[htbp]
    \centering
    \caption{Quantitative visual results on NVOS dataset.}
    \label{tab:nvos2}
    \small
    \begin{tabular}{l @{\hskip 10pt} c @{\hskip 5pt} c @{\hskip 5pt} c}
        \toprule
        \multirow{3}{*}{Method} & \multicolumn{3}{c}{CLIP-IQA~\cite{wang2022exploring} (\%) $\uparrow$} \\ \cmidrule(lr){2-4}
                                      & \scriptsize{\begin{tabular}[c]{@{}c@{}}\textit{Clear} / \textit{Unclear} \\ \textit{Boundary}\end{tabular}} & \scriptsize{\begin{tabular}[c]{@{}c@{}}\textit{Smooth} / \textit{Noisy} \\ \textit{Boundary}\end{tabular}} & \scriptsize{\begin{tabular}[c]{@{}c@{}}\textit{Complete} / \textit{Mutilated} \\ \textit{Object}\end{tabular}} \\ \midrule
        SAGD~\cite{hu2024semantic}      & 0.621 & 0.631 & 0.788 \\
        SA3D-GS~\cite{cen2023segment}      & 0.658 & 0.718 & 0.835 \\
        FlashSplat~\cite{flashsplat} & 0.626 & 0.644 & 0.829 \\
        COB-GS (ours)    & \textbf{0.682} & \textbf{0.731} & \textbf{0.859} \\ \bottomrule
    \end{tabular}
    \vspace{-10pt}
\end{table}

\begin{figure*}
    \centering
    \includegraphics[width=1.0\linewidth]{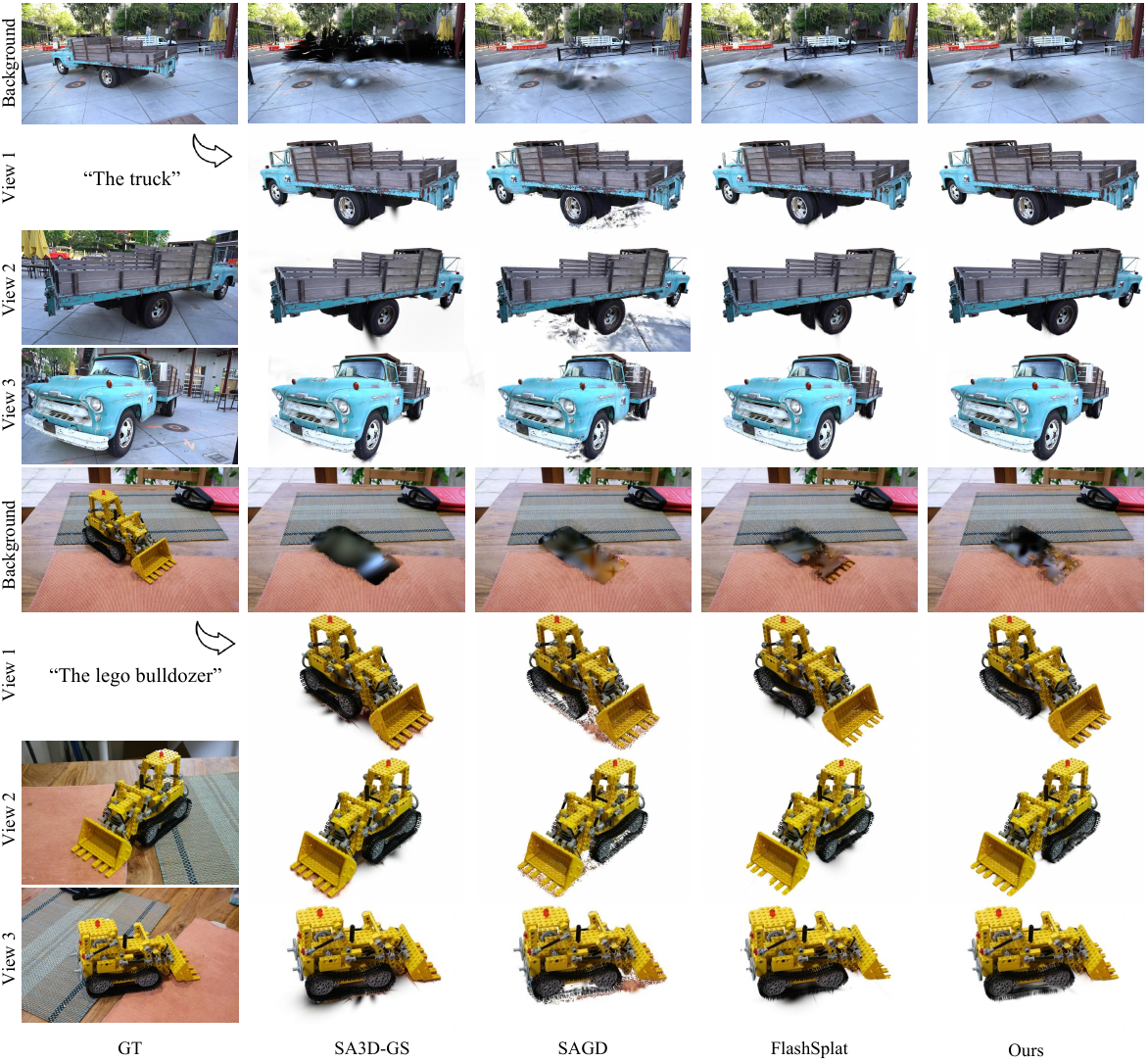}
    \vspace{-15pt}
    \captionsetup{singlelinecheck=false}
    \caption{
        Qualitative result of single-object segmentation. 
        The results show that our method segments the boundaries of the object more clearly, without blurred Gaussians, and the background is cleaner after the object removal.
    }
    \label{fig:visualization1}
\end{figure*}

\subsection{Qualitative Results}

We visualize the segmented objects and the background after object removal, including single-object and multi-object segmentation. We compared our results with the current SOTA 3DGS segmentation methods across multiple scenes.

\begin{figure}
    \centering
    \includegraphics[width=1.0\linewidth]{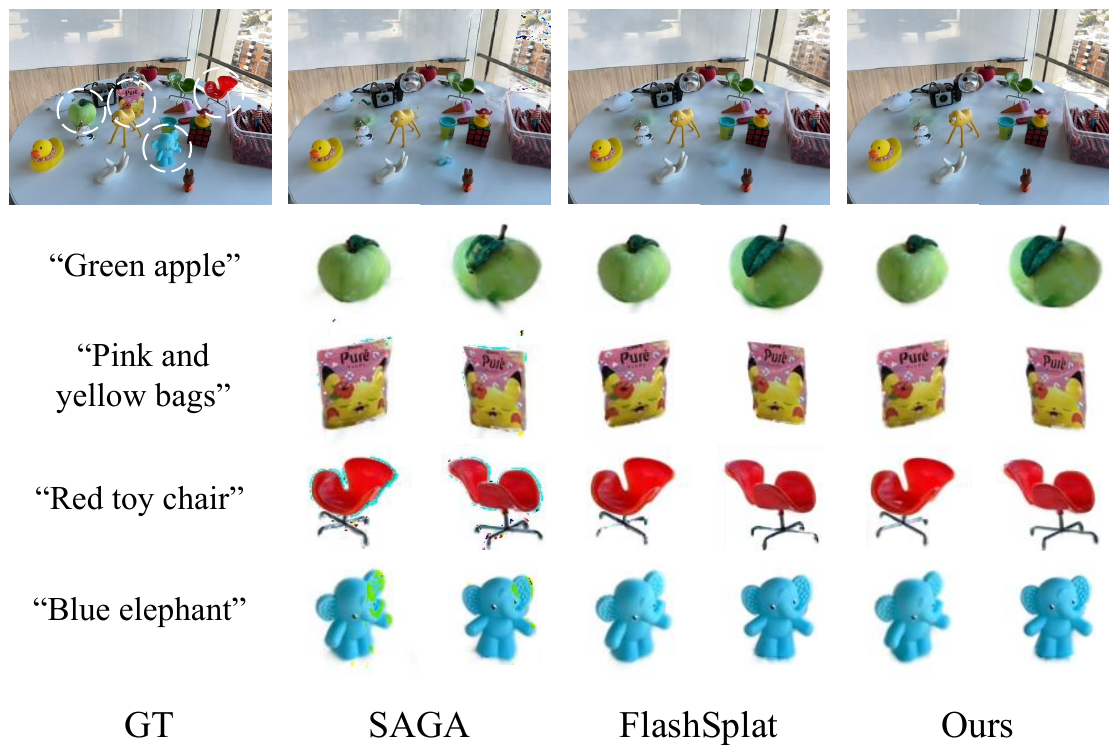}
    \vspace{-20pt}
    \captionsetup{singlelinecheck=false}
    \caption{
        Qualitative result of multi-object segmentation. 
        Our method gets more accurate segmentation results and clearer background quality after object removal compared to contrast method.
    }
    \label{fig:visualization2}
\end{figure}

For single-object segmentation, we selected the Truck scene from the T\&T dataset and the Kitchen scene from the MIP-360 dataset, as shown in Figure \ref{fig:visualization1}. To ensure fair comparison, consistent masks were used across all methods. We rendered the backgrounds and views of the segmented objects on different methods. 
In contrast, SA3D-GS~\cite{cen2023segment} introduces noticeable Gaussian blur at object boundaries, and its unbalanced loss function degrades the background quality. SAGD~\cite{hu2024semantic} leads to tiny edge Gaussians in the foreground. FlashSplat~\cite{flashsplat} reduces edge Gaussian blur by increasing background bias but indiscriminately removes semantically ambiguous real object regions, such as the truck’s rearview mirror and the Lego bucket in the Truck scene. Our method improves boundary representation accuracy while minimizing background distortion after object removal. Visualization confirms the visual quality assessment results in Table \ref{tab:nvos2}.

We further demonstrate the multi-object segmentation capability, using the Figurines scene from the LERF dataset, as illustrated in Figure \ref{fig:visualization2}. For segmentation, we employed masks based on text prompts consistent with the mask-based method FlashSplat~\cite{flashsplat}, while the feature-based method SAGA~\cite{cen2023saga} relies on point prompts. Both methods exhibit unclear boundaries and blurred backgrounds. In contrast, our approach achieves significantly clearer delineation of object boundaries while preserving background clarity.

\subsection{Ablation Study}
\subsubsection{Joint Optimization of Semantics and Textures}
An intuitive understanding is that relying solely on Gaussian splitting is insufficient. Although this method enables the Gaussian to adapt to object boundaries, it compromises visual quality. We conducted ablation experiments on the NVOS dataset to explore the relationship between semantics and texture, and the results are shown in Table \ref{tab:a1}.

\begin{table}[htbp]
    \centering
    \caption{Texture quality results on NVOS dataset (PSNR). ``Vanilla'' indicates the original scene; ``M.O'' indicates mask-optimized; ``T.O'' indicates texture-optimized.}
    \label{tab:a1}
    \vspace{-5pt}
    \renewcommand{\arraystretch}{1.4} 
    \setlength{\tabcolsep}{3pt} 
    {\fontsize{16}{16}\selectfont 
    \resizebox{\columnwidth}{!}{
        \begin{tabular}{@{\hskip 2pt}l@{\hskip 1pt}|cccccccc|c@{\hskip 1pt}} 
        \toprule
        Method & Fern & Flower & Fortress & Horns\_C & Horns\_L & Leaves & Orchids & Trex & Mean \\
        \midrule
        Vanilla & 24.26 & 26.75 & 29.43 & 22.24 & 22.24 & 15.07 & 19.82 & 24.68 & 23.06 \\
        M.O & 23.66 & 26.49 & 29.05 & 21.31 & 22.13 & 15.05 & 19.80 & 23.35 & 22.61 \\
        T.O & 24.26 & 26.69 & 29.45 & 22.27 & 22.27 & 15.06 & 19.83 & 24.71 & 23.07 \\
        M.O+T.O & 24.29 & 26.82 & 29.48 & 22.24 & 22.25 & 15.09 & 19.91 & 24.97 & 23.13 \\
        \bottomrule
        \end{tabular}
    }}
    \vspace{-20pt}
\end{table}

\begin{figure}
    \centering
    \includegraphics[width=1.0\linewidth]{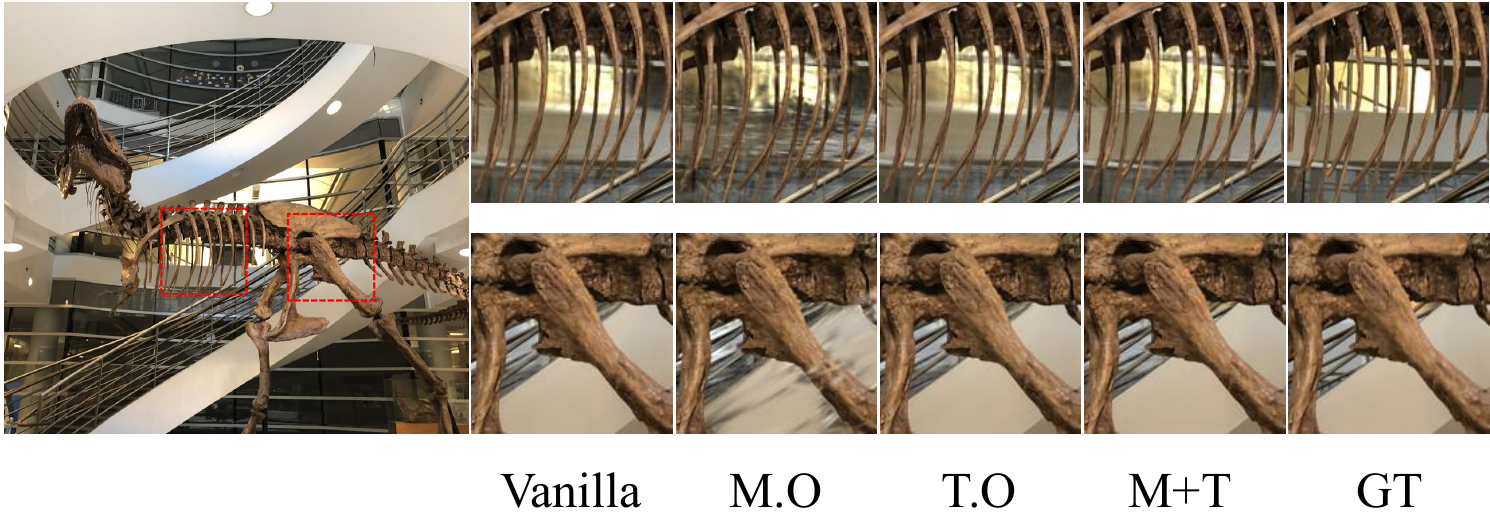}
    \vspace{-20pt}
    \captionsetup{singlelinecheck=false}
    \caption{
        Visualization results in the Trex scene. 
        ``Vanilla'' indicates the original result; ``M.O'' indicates mask-optimized; ``T.O'' indicates texture-optimized; ``M+T'' indicates M.O and T.O.
    }
    \label{fig:visualization3}
    \vspace{-5pt}
\end{figure}

\begin{table}[t]
    \centering
    \caption{Ablation results on NVOS dataset. BAGS indicates the boundary-adaptive Gaussian splitting; BGTR indicates the boundary-guided texture restoration; RAEM indicates robustness against erroneous masks.}
    \label{tab:a2}
    \renewcommand{\arraystretch}{0.9}
    \begin{tabular}{ccc|cc}
    \toprule
    \multicolumn{3}{c|}{Component} & \multicolumn{2}{c}{Performance} \\
    \midrule
    BAGS & BGTR & RAEM & mIoU (\%) & mAcc (\%)  \\
    \midrule
    &  &  & 91.2 & 98.3 \\
    \Checkmark &  &  & 91.9 & 98.5 \\
    \Checkmark & \Checkmark &  & 91.9 & 98.4 \\
    \Checkmark & \Checkmark & \Checkmark & \textbf{92.1} & \textbf{98.6} \\
    \bottomrule
    \end{tabular}
    \vspace{-10pt}
\end{table}

When only the boundary-adaptive Gaussian splitting is applied, scene texture quality significantly degrades. In contrast, joint optimization of mask labels and textures improves scene segmentation accuracy while preserving scene quality. For comparison, we performed the same number of iterations using texture optimization alone, which was less effective. Visualization results in Figure \ref{fig:visualization3} demonstrate that boundary Gaussian splitting based on mask label statistics can improve texture quality at object boundaries while maintaining independence between the foreground and background.

\subsubsection{Robustness Against Erroneous Masks}
After joint optimization, tiny ambiguous Gaussians remain along the segmentation boundaries due to predicted erroneous masks, as illustrated in Figure \ref{fig:fine gs}.
To investigate this, we conducted ablation experiments on the NVOS dataset, as shown in Table \ref{tab:a2}. The results indicate that without Gaussian splitting, large ambiguous Gaussians yield the lowest metrics. Applying only Gaussian splitting improves segmentation metrics but compromises texture quality.
Joint optimization improves texture quality at the expense of slightly decreasing segmentation accuracy. Finally, incorporating robust handling of tiny Gaussians gets optimal performance.

\begin{figure}
    \centering
    \includegraphics[width=1.0\linewidth]{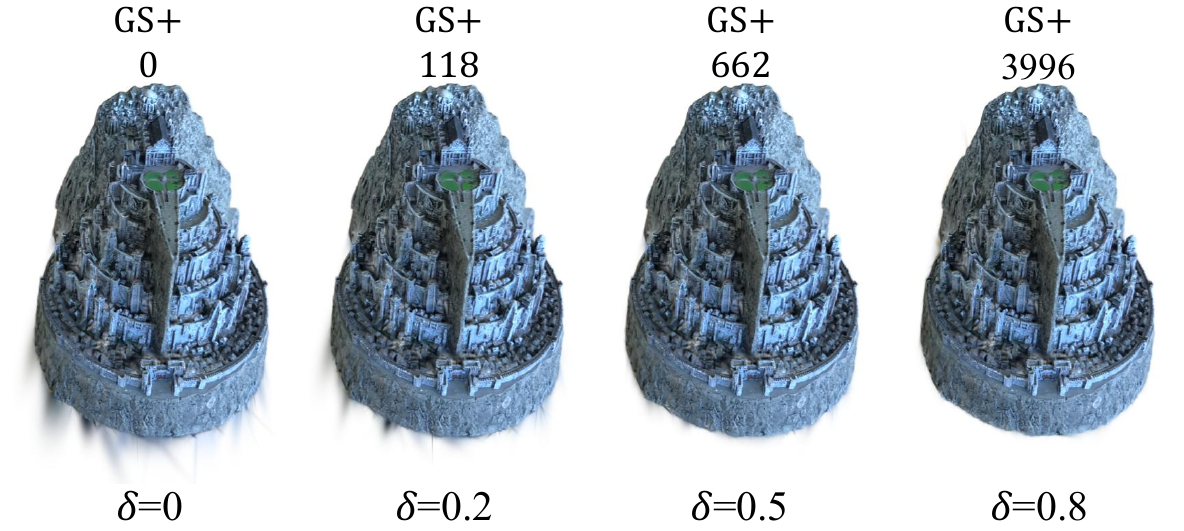}
    \vspace{-10pt}
    \captionsetup{singlelinecheck=false}
    \caption{
        Ablation results in the Fortress scene. 
As $\delta$ increases, the number of ambiguous boundary Gaussians decreases; however, the total number of Gaussians increases rapidly. 
    }
    \label{fig:visualization4}
    \vspace{-10pt}
\end{figure}

\subsubsection{Ambiguity Supervision Threshold}
In our method, the parameter $\delta$ functions as a threshold to control the number of ambiguous Gaussians. As $\delta$ increases, the discrimination against semantically ambiguous Gaussians strengthens, leading to more splits. We conducted an ablation study on $\delta$ in the Fortress scene, as shown in Figure \ref{fig:visualization4}.
We observed that increasing $\delta$ reduces ambiguous boundary Gaussians and improves segmentation clarity. However, it also introduces a substantial number of additional Gaussians at the object edge. Therefore, selecting an optimal $\delta$ is crucial, with $\delta = 0.5$ proving effective for most scenarios.

%% file: sec/5_discussion_and_conclusion.tex
\section{Conclusion}
\label{discussion and conclusion}

In this paper, we propose COB-GS, a 3DGS refinement and segmentation approach that clearly segments scene boundaries. COB-GS is innovatively designed to jointly optimize semantics and textures, ensuring they complement each other. This process is supported by a boundary-adaptive Gaussian splitting method. Specifically, in the semantic optimization phase, we utilize semantic gradient statistics to identify and split the ambiguous Gaussians, aligning them with object boundaries. Then we enhance the scene texture along the precisely refined boundary structures. Experimental results demonstrate significant improvements in 3D segmentation performance, particularly in terms of clear object boundaries, accurate textures and robustness against inaccurate masks. In summary, COB-GS is the first 3DGS segmentation method explicitly designed to jointly optimize semantics and texture, ensuring they enhance one another and offering new insights into segmentation with learnable scene representations. Currently, 3D scene reconstruction encounters the challenge of floating artifacts, which are magnified after the segmentation. Future work should focus on effectively eliminating these floating artifacts using semantic information.

\vspace{3pt}


%% file: sec/X_suppl.tex
\clearpage
\maketitlesupplementary
In the supplementary material, we first introduce in detail our proposed two-stage mask generation based on text prompts in \cref{method:Two-Stage Mask Generation}. Next, we present the concrete training strategy and implementation details of COB-GS in \cref{exp:Implementation Details}. In \cref{exp:OV3DS} and \cref{exp:Computation Cost}, we  evaluate the open-vocabulary segmentation capability and the computational cost of COB-GS, respectively. In \cref{re:Application Scenarios} and \cref{re:Multi-granularity Segmentation}, we discussed the differences between mask-based and feature-based methods in terms of application scenarios and multi-granularity segmentation. Finally, additional visualizations of the segmentation results are presented in \cref{exp:More Quantitative Results}.
\section{Two-Stage Mask Generation Based on Text Prompts}
\label{method:Two-Stage Mask Generation}

Mask-based 3D segmentation requires generating a set of masks for regions of interest from a collection of input images. Thus, the supervision data consists of \( V \) input views \( \{I^v\} \) corresponding to 2D binary masks \( \{M^v\} \). Each mask \( M \in \mathbb{R}^{H \times W} \) contains discrete values of 0 and 1. The related work SA3D~\cite{cen2023segment} improves optimization efficiency and mask view consistency by using Segment Anything Model (SAM)~\cite{kirillov2023segany} to iteratively generate the mask for each frame. With the emergence of foundational models like SAM2~\cite{ravi2024sam2}, mask prediction across video sequences has become feasible.

SAM2 retains the encoder-decoder structure of SAM, where the encoder \( S_e \) takes an image \( I \) as input. Unlike SAM, SAM2 employs memory attention \( S_m \) to tilize past frame features \( f_m \) as conditions for generating the current frame embedding \( e_I \):

\[
e_I = S_m(S_e(I), f_m) \tag{1}
\]

The past frame features \( f_m \) are maintained in a FIFO memory queue. The decoder takes the current frame embedding \( e_I \) and the prompts \( P \) as input, outputting the corresponding 2D binary mask \( M \):

\[
M = S_d(e_I, P) \tag{2}
\]

The prompts \( P \) include masks, boxes, points, or texts. The memory capability of SAM2 allows it to handle mask prediction for video sequences, which aligns with the input view conditions \( \{I^v\} \) for the 3DGS task. However, when SAM2 performs mask prediction across video sequences, it encounters challenges with object continuity; specifically, it may fail to recognize severely occluded objects due to information discontinuity.To address this issue, we propose a two-stage mask generation method based on text prompts. In the coarse mask generation stage, we utilize Grounding DINO~\cite{caron2021emerging}  to extract box prompts from the given prompt frame with lower text confidence, which are then used for full-sequence mask prediction to obtain preliminary results. In the fine-grained stage, we leverage Grounding DINO with higher text confidence to extract box prompts for subsequences within the original sequence that lack mask prediction results, which are then used for subsequence mask prediction. See \cref{alg:Two-stage mask generation} for details.

\begin{algorithm}[t]
    \caption{Two-stage mask generation}
    \label{alg:Two-stage mask generation}
    \begin{spacing}{1.05}
    \begin{algorithmic}
        \STATE \textbf{Input}: Frame index $idx$, text prompt $text$, image set $I$, high confidence $C_{high}$, low confidence $C_{low}$
        \STATE \textbf{Result}: Updated dictionary $video\_segments$
        \STATE Initialize dictionary $valid\_idxs \gets \{\}$
        \STATE Initialize dictionary $video\_segments \gets \{\}$
        
        \STATE SAM2.init\_state($I$) 
        \STATE $image \gets I[idx]$ 
        
        \STATE $boxes \gets \text{Grounding DINO}(text, image, C_{low})$ 
        
        \STATE SAM2.add\_new\_box($idx,boxes$) 

        \FOR{each frame $i,mask$ in SAM2($idx$)}
            \STATE $video\_segments[i] \gets mask$ 
            \STATE $valid\_idxs[i] \gets \text{if } mask \text{ is empty then } 0 \text{ else } 1$
        \ENDFOR
        
        \FOR{each $key$ in $valid\_idxs$}
            \IF{$valid\_idxs[key] = 0$}
                \STATE $boxes \gets \text{Grounding DINO}(text, I[key], C_{high})$  
                
                \IF{boxes is empty}
                    \STATE \textbf{continue} 
                \ENDIF
                
                \STATE SAM2.add\_new\_box($idx,boxes$) 
                \STATE $max\_sk \gets \text{FindMaxSub}(valid\_idxs, key)$  
                
                \FOR{each frame $j, mask$ in SAM2($key,max\_sk$)}
                    \STATE $video\_segments[j] \gets mask$  
                    \STATE $valid\_idxs[j] \gets \text{if } mask \text{ is empty then } 0 \text{ else } 1$
                \ENDFOR

            \ENDIF
        \ENDFOR
    \end{algorithmic}
    \end{spacing}
\end{algorithm}
\vspace{-5pt}
\section{Implementation Details}
\label{exp:Implementation Details}
\vspace{-5pt}
\begin{table*}[htbp]
    \centering
    \caption{Results on LERF-mask dataset.}
    \label{tab:lerf}
    \begin{tabular}{lcccccc}
        \toprule
        Method               & \multicolumn{2}{c}{Figurines} & \multicolumn{2}{c}{Ramen} & \multicolumn{2}{c}{Teatime} \\ 
        \cmidrule(lr){2-3} \cmidrule(lr){4-5} \cmidrule(lr){6-7}
                           & mIoU (\%) & mBIoU (\%) & mIoU (\%) & mBIoU (\%) & mIoU (\%) & mBIoU (\%) \\ 
        \midrule
        DEVA ~\cite{cheng2023tracking}              & 46.2      & 45.1       & 56.8      & 51.1       & 54.3      & 52.2       \\
        LERF ~\cite{lerf2023}               & 33.5      & 30.6       & 28.3      & 14.7       & 49.7      & 42.6       \\
        SA3D ~\cite{cen2023segment}              & 24.9      & 23.9       & 7.4       & 7.0        & 42.5      & 39.2       \\
        LangSplat ~\cite{qin2023langsplat}         & 52.8      & 50.5       & 50.4      & 44.7       & 69.5      & 65.6       \\
        Gaussian Grouping ~\cite{gaussian_grouping}  & 69.7      & 67.9       & 77.0      & 68.7       & 71.7      & 66.1       \\
        COB-GS (ours)    & \textbf{76.3} & \textbf{73.9} & \textbf{78.1} & \textbf{69.2} & \textbf{77.2} & \textbf{72.8} \\ 
        \bottomrule
    \end{tabular}
\end{table*}

\begin{figure*}
    \centering
    \includegraphics[width=1.0\linewidth]{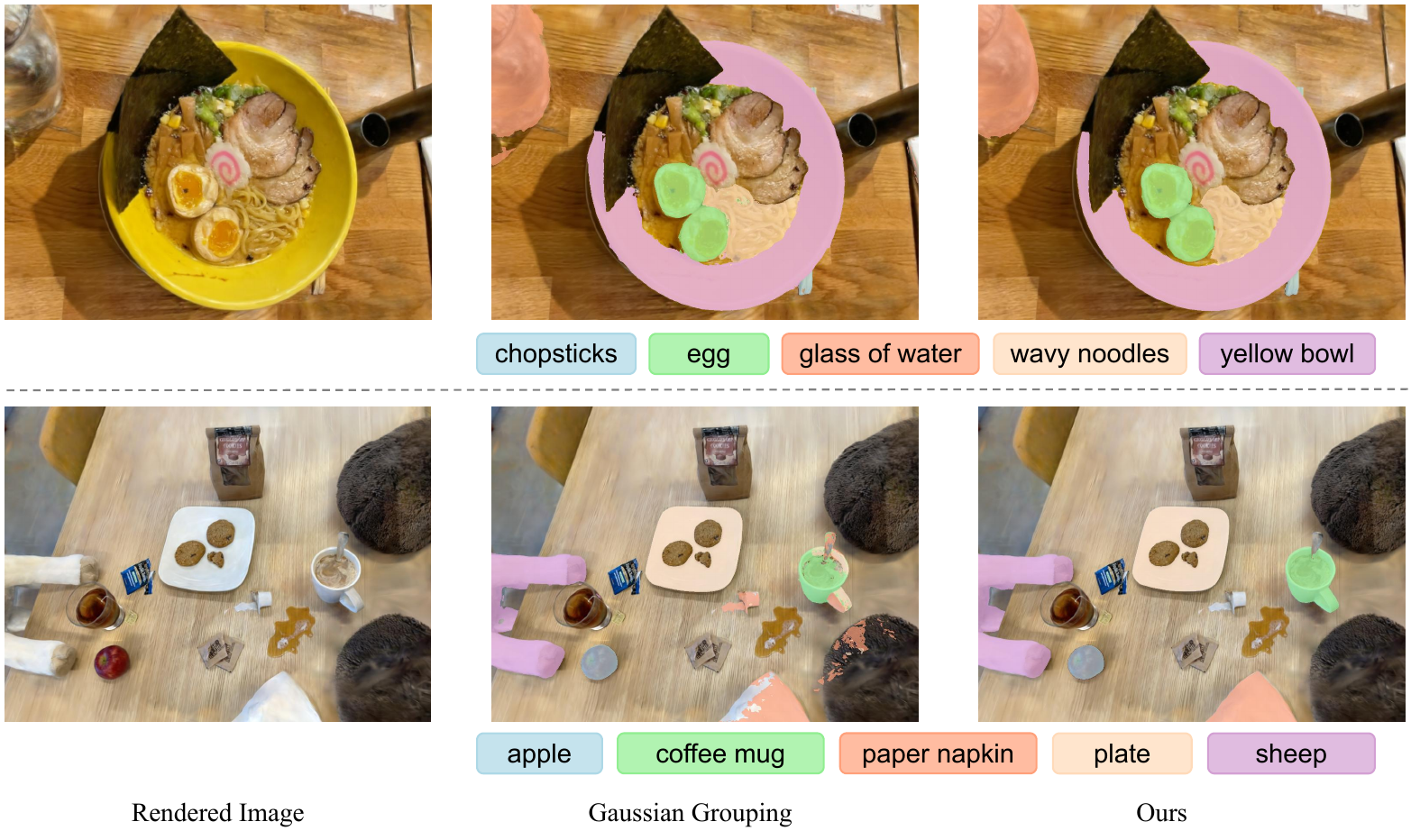}
    \captionsetup{singlelinecheck=false}
    \vspace{-15pt}
    \caption{
        Visualization of the LERF-mask dataset~\cite{gaussian_grouping}.
The result of the segmentation is obtained under the specified text prompt. 
    }
    \label{fig:supp1.png}
    \vspace{-12pt}
\end{figure*}

Our method is a post-processing method based on the original 3D Gaussian Splatting~\cite{kerbl20233d}. For each scene, we perform 30,000 iterations of training according to the parameters set by the original 3DGS to obtain the original 3DGS scene. COB-GS mainly consists of two components: optimization process and robustness process. The optimization process involves alternating between mask optimization and texture optimization. For the mask optimization stage, we optimize the mask labels and perform Gaussian splitting. The learning rate of the mask labels is set to 0.1. For the texture optimization stage, we optimize the geometry and texture, and the learning rate of appearance follows the original 3DGS setting. Each stage is trained for 2×$V$ iterations, where $V$ is the number of input images. Two sets of hyperparameters are used for different scene types: for forward scenes, we set $\delta = 0.5$ and perform a total of 22×$V$ iterations of alternating optimization; for surrounding scenes, we set $\delta = 0.8$ and conduct 14×$V$ iterations of alternating optimization. The robustness process follows scene optimization and involves extracting and refining ambiguous boundary Gaussians at scales smaller than the pixel scale. In our two-stage mask generation method, we utilize the SAM2 hiera\_l model and the Grounding DINO swinb model. All experiments were conducted on a single NVIDIA RTX 3090 GPU.

\begin{figure*}
    \centering
    \includegraphics[width=1.0\linewidth]{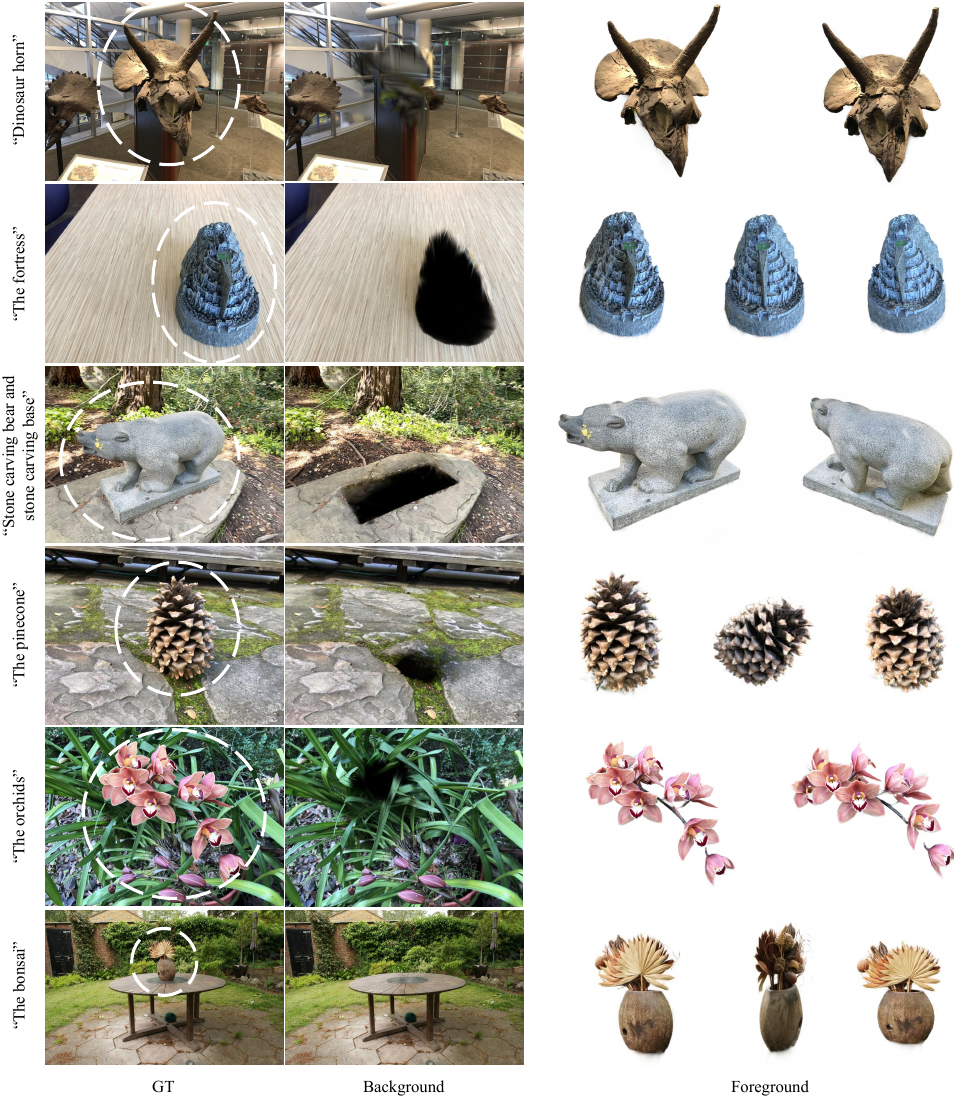}
    \captionsetup{singlelinecheck=false}
    \caption{
        Visualization of 3DGS segmentation.  
        We utilize text prompts to obtain object masks and perform 3D segmentation across multiple scenes, including Horns, Orchids and Fortress from the LLFF dataset, Garden from MIP-360, Bear from the IN2N dataset, and Pinecone from NeRF.
    }
    \label{fig:supp2}
\end{figure*}

\section{Open-Vocabulary 3D Segmentation}
\label{exp:OV3DS}
To achieve open-vocabulary semantic segmentation, we follow the setup of existing methods~\cite{cen2023segment,gaussian_grouping} and utilize Grounding DINO~\cite{caron2021emerging} to generate boxes for input images, similar to the approach in \cref{method:Two-Stage Mask Generation}. We compare our method with the current state-of-the-art methods for open-vocabulary 3D segmentation using the LERF-mask dataset, which is annotated from test views of three 3D scenes in the LERF dataset. The scenes contain severe object occlusions, and the mask boundaries of the test views are more complex. As shown in Table \ref{tab:lerf}, our method demonstrates a clear advantage over current SOTA methods. Visual segmentation comparisons in Figure \ref{fig:supp1.png} reveal that our method provides more accurate segmentation predictions with clear boundaries, while Gaussian Grouping~\cite{gaussian_grouping} exhibits blurriness in segmentation results.

\section{Computation Cost}
\label{exp:Computation Cost}

We evaluate the computational efficiency of COB-GS in comparison to state-of-the-art 3DGS segmentation methods, namely the feature-based SAGA~\cite{cen2023saga} and the mask-based FlashSplat~\cite{flashsplat}. This evaluation is conducted on the Fortress scene ($V=42$) from the LLFF dataset~\cite{mildenhall2019llff} using a single NVIDIA RTX 3090 GPU, with results presented in Table \ref{tab:cost}. 
We provide the total time cost (prep time+opt time+seg time) and the maximum VRAM of the entire reconstruction and segmentation pipeline.
SAGA~\cite{cen2023saga} requires 10,000 iterations of gradient descent to distill 2D masks into object features associated with each 3D Gaussian, resulting in substantial additional training time for scene optimization. Moreover, object segmentation remains time-consuming due to the need for network inference. FlashSplat~\cite{flashsplat} does not offer a mask extraction method, and assigning labels to each Gaussian through forward rendering is relatively time-consuming. In contrast, our extraction process relies entirely on inverse rendering, which ensures that texture optimization simultaneously optimizes scene labels. The optimization time is comparable to the speed of FlashSplat, and segmentation requires only filtering the labels.

\begin{table}[htbp]
    \centering
    \small
    \renewcommand{\arraystretch}{0.9}
    \setlength{\tabcolsep}{4pt} 
    \begin{tabular}{@{}l c c c c c@{}}
        \toprule
        Method      & \begin{tabular}[c]{@{}c@{}}Prep \\ Time\end{tabular} & \begin{tabular}[c]{@{}c@{}}Opt \\ Time\end{tabular} & \begin{tabular}[c]{@{}c@{}}Seg \\ Time\end{tabular} & \begin{tabular}[c]{@{}c@{}}Total \\ Time\end{tabular} & Mem \\ \midrule
        SAGA~\cite{cen2023saga}        & 145 s              & 20 min         & 200 ms         & 22.42 min            & 7.6 G \\
        FlashSplat~\cite{flashsplat}  & N/A                  & 24 s           & 10 ms         & N/A             & 2.4 G \\
        COB-GS    & 4 s       & 24 s           & 8 ms         & 0.46 min              & 2.7 G \\
        \bottomrule
    \end{tabular}
    \caption{Computation cost comparisons over the Fortress scene.}
    \label{tab:cost}
\end{table}

\section{Application Scenarios}
\label{re:Application Scenarios}
\cref{tab:cost} shows that feature-based methods like SAGA~\cite{cen2023saga} consume more time and memory than mask-based methods. This is because feature-based methods optimize high-dimensional features for the entire scene, while our approach focuses solely on optimizing labels. This difference is evident in the optimization time. However, SAGA~\cite{cen2023saga} has the advantage of allowing multiple segmentations with a single training session, making it suitable for offline fixed scenes that require frequent segmentations, despite its high equipment demands.In contrast, mask-based methods assign labels directly on the reconstructed 3DGS, offering faster single segmentation. COB-GS achieves better visual quality in an negligible time and suits edge applications that require fast and detailed single object segmentation. Furthermore, progress in foundational models and faster 3DGS training will further promote COB-GS's real-time applicability.

\section{Multi-granularity Segmentation}
\label{re:Multi-granularity Segmentation}
Unlike the feature-based method N2F2~\cite{N2F2}, which requires a time-consuming process to integrate multi-granular high-dimensional features for scene reconstruction, the mask-based COB-GS decouples the processes of reconstruction and segmentation. This separation allows for greater flexibility, enabling arbitrary granularity to be finely achieved during the mask generation phase. As a result, COB-GS an obtain the region of interest efficiently within the preparation time of a single fast segmentation, streamlining the workflow and enhancing performance without the heavy computational burden associated with integrating complex features.

\section{More Qualitative Results}
\label{exp:More Quantitative Results}
To demonstrate the effectiveness of our proposed 3D segmentation method in producing clear object boundaries, we provide visualizations of 3D segmentation across multiple scenes, including the Horns, Orchids and Fortress from the LLFF dataset~\cite{mildenhall2019llff}, the Garden from MIP-360~\cite{barron2021mipnerf}, the Bear from the IN2N dataset~\cite{instructnerf2023}, and the Pinecone from NeRF~\cite{mildenhall2020nerf}, encompassing both forward and surrounding scenes. We obtain masks using text prompts, as described in \cref{method:Two-Stage Mask Generation}. The results shown in Figure \ref{fig:supp2} clearly demonstrate that the object edges in our 3D segmentation results are very clear, while also maintaining high-quality textures for both the foreground and background.